\documentclass{article}

\usepackage[final]{corl_2020} 
\usepackage[pdftex]{graphicx}
\usepackage{amsmath}
\usepackage{amssymb}
\usepackage{color}
\usepackage{wrapfig}
\usepackage{multirow}

\newcommand{\ch}[1]{#1}
\newcommand{\chch}[1]{{#1}}

\title{Modeling Long-horizon Tasks \\ as Sequential Interaction Landscapes}

\author{
  S\"{o}ren Pirk\hspace{11mm}Karol Hausman\hspace{11mm}Alexander Toshev\\
  Google AI, Robotics at Google\\
  \texttt{\{pirk, karolhausman, toshev\}@google.com} \\
   \And
   Mohi Khansari \\
   X \\
   \texttt{khansari@x.team} \\
}

\begin{document}
\maketitle


\begin{abstract}
Task planning over long time horizons is a challenging and open problem in robotics and its complexity grows exponentially with an increasing number of subtasks. In this paper we present a deep neural network that learns dependencies and transitions across subtasks solely from a set of demonstration videos. We represent each subtasks as \textit{action symbols}, and show that these symbols can be learned and predicted directly from image observations. Learning symbol sequences provides our network with additional information about the most frequent transitions and relevant dependencies between subtasks and thereby structures tasks over long-time horizons. Learning from images, on the other hand, allows the network to continuously monitor the task progress and thus to interactively adapt to changes in the environment. We evaluate our framework on two long horizon tasks: (1) block stacking of puzzle pieces being executed by humans, and (2) a robot manipulation task involving pick and place of objects and sliding a cabinet door with a 7-DoF robot arm. We show that complex plans can be carried out when executing the robotic task and the robot can interactively adapt to changes in the environment and recover from failure cases.
A video illustrating live-action captures of our system is provided as supplementary material.
\end{abstract}



\section{INTRODUCTION}

Enabled by advances in sensing and control, robots are getting more capable of performing intricate tasks in a robust and reliable manner. In recent years, learned policies for control in robotics have shown impressive results~\cite{finn2017deep}. However, learning a single black-box function mapping from pixels to controls remains challenging.
In particular, complex manipulation tasks require to operate on diverse sets of objects, their locations, and how they are manipulated. Simultaneously reasoning for both, the 'what' (e.g. which object) and the 'how' (e.g. grasping), is a challenging problem. Additionally, due to the long time horizon of many tasks, the model can only observe a small portion of the full task at any given time. This partial observability increases with longer tasks and higher complexity.

Current learning-based planning approaches either focus on object representations~\cite{janner2018reasoning}, on learning sequences of symbols without rooting the plans in the actual environment~\cite{doi:10.1111/cgf.13644}, or generate plans based on explicit geometric representations of the environment~\cite{doi:10.1177/0278364908097884,5509563}. Formulating plans without feedback from the environment does not easily generalize to new scenes and is inevitably limited to static object arrangements. Generating plans based on image data, e.g. by predicting future images, limits the planning horizon to only a few steps~\cite{finn2017deep}. More recently, reinforcement learning approaches have shown initial success in solving robot manipulation tasks~\cite{qtopt,openai2019dexterous}, however, the end-to-end learning of long-horizon, sequential tasks remains challenging. 

\begin{wrapfigure}{r}{0.55\textwidth}
  \begin{center}
  \includegraphics[width=1.0\linewidth]{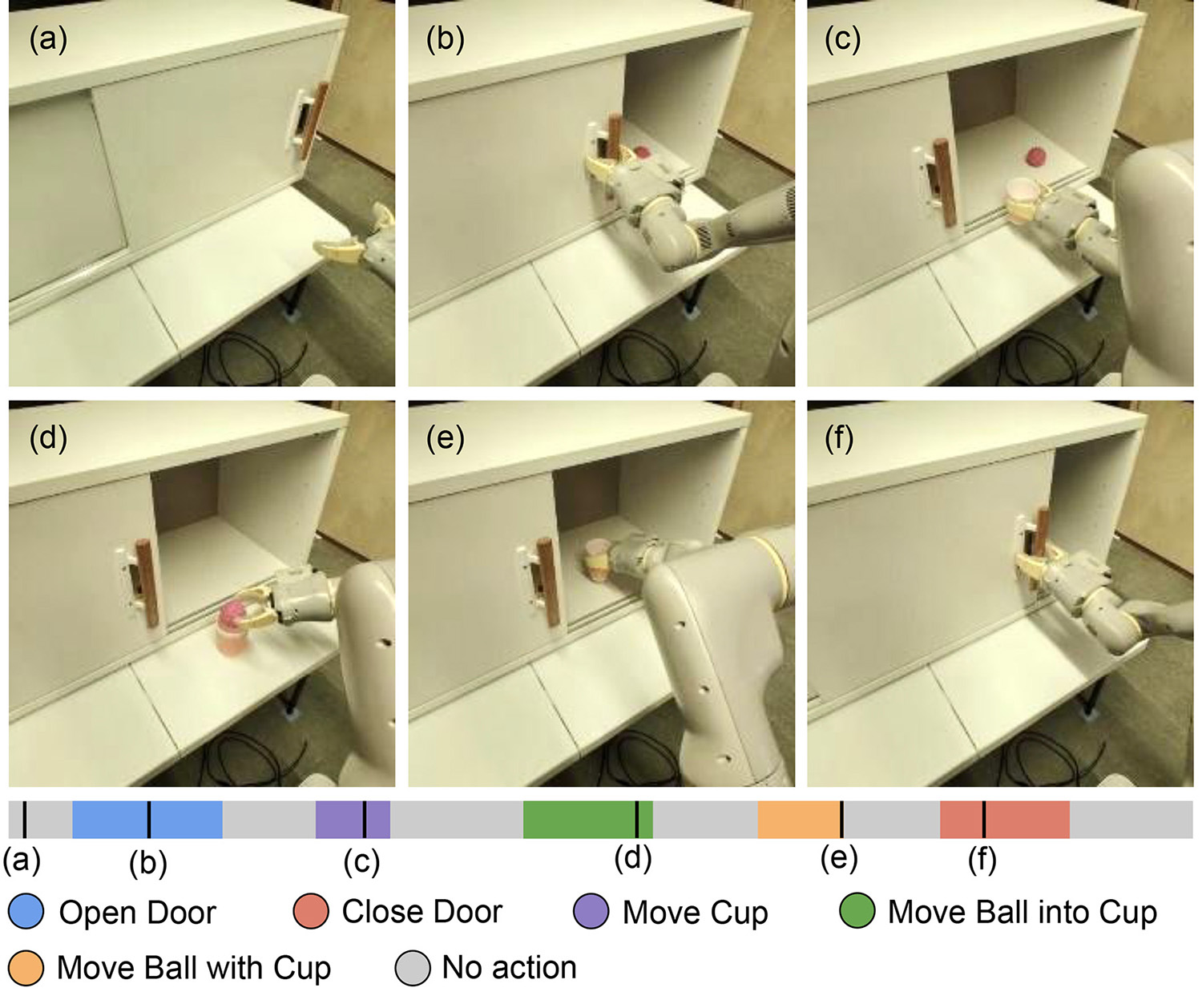} 
  \end{center}
  \vspace{-3mm}
  \caption{Robot performing a long-horizon manipulation task of re-arranging objects inside a cabinet. We decompose these tasks into a sequence of abstract actions, e.~g.~`close door', `move cup', etc. Thus, a task spanning over hundreds of image frames can be compactly summarized as a sequence of abstract symbols, which (1) can be accurately predicted at execution time; and (2) precisely executed by a robot.
  }
  \label{fig:seq_cabinet_robot}   
  \vspace{-3mm}
\end{wrapfigure}

In this paper, we propose a two layer representation of complex tasks, where we use as an intermediate representation a set of abstract actions or sub-tasks (see Fig.~\ref{fig:seq_cabinet_robot}).
Each action is represented by a symbol that describes what needs to happen to complete a sub-task in an abstract manner (e.g. move cup). This discretization allows us to reason about the structure of tasks without being faced with the intricacies of real environments and the related physics (e.g. object pose).

\setlength{\intextsep}{5pt}%
\setlength{\columnsep}{10pt}%

Each symbol is then used to select an individual policy that describes how an object or the agent itself need to be manipulated toward the higher-level goal. When executing an action we can then consider the complexity imposed by the real scene, such as finding an object or identifying its pose to grasp it. Our goal is to execute complex and long-horizon tasks by learning the sequential dependencies between task-relevant actions. To learn sequences of sub-tasks while respecting changes in the scene, we employ a sequence-to-sequence model, commonly used for natural language processing, to translate sequences of image embeddings to action symbols~\cite{Sutskever:2014:SSL:2969033.2969173,Cho2014LearningPR}. 

We test the capabilities of sequence prediction by evaluating our framework on two environments. First, we use a robot arm to manipulate objects in an office environment, where the goal is to find objects in a cabinet, perform operations on the objects, and to move them back into the cabinet. In the environment shown in Fig.~\ref{fig:seq_cabinet_robot} the task is to find a cup, to put a ball in the cup, and to move both objects together back to the cabinet. Different sequences of sub-tasks can lead to a successful completion of the task. For example, while the robot has to first open the cabinet door, it can then either move the cup or the ball outside the cabinet, to eventually put the ball in the cup and both objects back into the cabinet. For the second experiment, we perform a stacking task that requires to move blocks from an initially random configuration to three stacks of blocks. 

We evaluate and discuss the success of these experiment and demonstrate that using action symbols allows us to organize tasks as sub-tasks. We empirically evaluate our model, both in an off-line and on-line fashion on two manipulation tasks. In summary, our contributions are: (1) we propose a deep learning approach that learns dependencies and transitions across subtasks as action symbols from demonstration videos; (2) we introduce a framework that integrates our proposed approach into existing state-of-the-art robotics work on motion primitives \cite{schaal05}; (3) we embed IMU sensors into objects to track object motion and in turn to automatically obtain action symbols; (4)~we evaluate the learned sequence model on two long-horizon tasks, showing that sequences of action symbols can be predicted directly from image observations and be executed in a closed-loop setting by a robot.

\vspace{-3mm}
\section{RELATED WORK}
\vspace{-3mm}
The topic of long-term planning has received a considerable amount of attention in the past. Here we provide an overview of related work with a focus on planning and learning-based approaches.

\vspace{-1mm}
\textbf{Motion and Manipulation Planning:} traditionally, approaches for planning have focused on computing trajectories for robotic motion to, for example, arrange  objects in specified arrangements~\cite{1087847}. Many approaches jointly solve for planning tasks and motion to enable more informed robotic behavior. Examples include: focusing on the explicitly modeling of geometric constraints as part of task planning~\cite{Lagriffoul:2014:ECT:2693308.2693309}, leveraging physics-based heuristics~\cite{7733599}, hierarchical planning~\cite{5980391,1087847}, probabilistic models~\cite{7139389} or integrating action parameters as part of learning forward models~\cite{7759573}. 
More recently, various approaches started to explicitly focus on planning with neural networks. As an example, Zhang et al.~\cite{zhang2018composable} propose to use user-defined attributes in environments to then learn policies that enable transitioning between these features of interest.  

\vspace{-1mm}
\textbf{Symbolic Planning and Movement Primitives:} it has been recognized that abstract symbols can serve as a meaningful representation to structure tasks~\cite{doi:10.1177/0278364908097884} and to organize their often hierarchical properties~\cite{5980391}. In the work of Ortehy et al.~\cite{6630974}, motion primitives -- represented as symbols -- are optimized on a geometric level to validate predictions. Garett et al.~\cite{doi:10.1177/0278364917739114} combine symbolic planning with heuristic search to efficiently perform tasks and motion planning while Muxfeldt et al.~\cite{7733742} and Xu et al.~\cite{xu2018} hierarchically  decompose task into assembly operations and sub-task specifications. 
\chch{Movement primitives are another well-established approach to represent basic movements (e.g. grasping) for completing tasks and a number of approaches exist that focus on the parallel activation and the smooth blending of movement primitives~\cite{schaal05,LioutikovKPM2014,Paraschos2018,doi:10.1177/0278364919868279,4650953}.
}

\vspace{-1mm}
\textbf{Learning Skills:} it has been shown that a single task can be represented by hand-crafted state machines, where motion primitives allow us to transition between individual states~\cite{1087032,FIKES1972251,Sen2016AutomatingMM}. While these approaches provide a powerful means to represent agent behavior, their specification needs to be adapted manually for individual tasks, which prevents their use at scale. 
To address this issue, there has a been a number of hierarchical imitation
learning approaches~\cite{intentionGAN, DDO, compile,directedinfoGAIL} that focus on segmenting long-horizon tasks into subcomponents. 
As an alternative, reinforcement learning approaches aim at learning policies in an end-to-end manner by obtaining task-relevant features from example data~\cite{mnih2015humanlevel}. Combined with deep neural networks, this has shown to be a promising direction to learn object assembly~\cite{8202244} or more advanced motion models~\cite{Amiranashvili2018MotionPI}. 
Another direction is to learn policies for robot-object interaction from demonstrations with the goal to reduce uncertainty with expert
demonstrations~\cite{Krishnan2017,niekum2014,Niekum2013IncrementalSG,robotics7020017}. Despite these promising efforts, learning policies from demonstrations that generalize to new tasks still is an open research problem.

\vspace{-1mm}
\textbf{Understanding agent-object interactions:} understanding object motion and object-agent interactions enables reliably learning agent behavior to manipulate objects. 
Object-centric representations, also learned from visual data, can serve as a powerful means to understand physical agent and object interactions~\cite{7298903,Devin2017DeepOR,9196567,doi:10.1111/cgf.13644,Pirk:2017:UEO:3087678.3083725}. As a recent example, Janner et al.~\cite{janner2018reasoning} propose an object-oriented prediction and planning approach to model physical object interactions for stacking tasks. However, obtaining meaningful signals from visual data is often difficult in real-world settings due to cluttered scenes and occlusions. 
\ch{Unlike existing work that mostly focuses on learning action symbols implicitly -- e.g. as latent variables -- we represent actions explicitly, which  provides more semantics of a task. Furthermore, we learn the action symbols directly from sequences of images.} This facilitates to infer the correct order of actions necessary to complete a task, while our method also allows us to respond to changes in the environment. Each individual action is then executed with an individual policy.
\vspace{-2mm}
\section{METHOD}
\vspace{-2mm}
Our main goal is to learn the sequential structure of tasks by factorizing them into task-relevant actions. This is motivated by the observation that many tasks are as well combinatorial as they are continuous. They are combinatorial in that an agent has to select among a discrete set of objects to perform a task. For example, a stacking task requires to arrange a specific number of objects. However, an agent has to also operate in a physical environment that requires to continuously interact with objects.

Optimizing for both of the aforementioned factors to perform long-term planning is challenging due to the uncertainty imposed by the actual scene. Therefore, to perform long-term planning, we first factorize long-horizon tasks into a discrete set of task-relevant actions. These actions represent what needs to happen to complete a sub-task, but at a very high-level of abstraction and without any notion of how an agent has to perform the action. For example, an action might just be `move cup'. We denote the combinatorial complexity of all possible actions to perform a task as \textit{interaction landscape}. Second, once a task is structured into task-relevant actions we use expert policies obtained from learned demonstrations to perform individual actions.

Similar to existing approaches~\cite{doi:10.1177/0278364908097884,5980391,6630974}, we propose to use a set of \textit{action symbols} as an abstract representation of sub-tasks. These symbols represent basic actions, such as `open door', `move cup', `put ball', etc., and are defined for different tasks (Supplementary Material, Table~\ref{table:action_symbols}). Sequences of symbols are intended to provide an abstraction of the task that can be learned to be predicted and then executed by a robot. The set of symbols is denoted as $K$. We use the action symbols to train an encoder-decoder sequence-to-sequence model, that translates sequences of image embeddings to sequences of action symbols. This allow us to predict the next action based on the current state of the scene as well as according to which sub-tasks are already completed.

\vspace{-2mm}
\subsection{Action-centric Representation}
\vspace{-2mm}

To obtain a representation of the scene as well as of the ongoing actions we use a pretrained ResNet50~\cite{He2015DeepRL}, with one additional last layer (16 dimensions) trained as classifer against per frame action symbols as labels. We use the resulting 16-dimensional embedding as a compact representation of image features for each frame of an input sequence. 

We then employ sequence models~\cite{Sutskever:2014:SSL:2969033.2969173,Cho2014LearningPR} to predict future action symbols given a history of image embeddings. Given a sequence of image embeddings $(E_1, \ldots, E_t)$ up to current time $t$, we predict the next $k$ action symbols $(a_{t+1}, \ldots, a_{t+k})$:
\[
a_{t+1}, \ldots, a_{t+k} = \textrm{SeqMod}(E_1, \ldots, E_t).
\]
We cast the above formulation as a `translation' of image embeddings to action symbol sequence. Therefore, we employ a sequence-to-sequence model~\cite{Sutskever:2014:SSL:2969033.2969173}, an established neural translation formulation, where we map the embedding sequence to an action sequence. In more detail, the sequence-to-sequence model consists of an encoder and decoder LSTM. The encoder consumes the input images as sequence of embeddings and encodes it into a single vector, which is subsequently decoded into an action symbol sequence by a second LSTM (Supplementary Material, Fig.~\ref{fig:encoder_decoder}). Using image embeddings as high-dimensional continuous inputs is one of the major differences to the original translation application of the above model.

Learning the sequential structure of tasks based on image embeddings and action symbols enables us to perform tasks in varying combinations of sub-tasks and depending on a given scene configuration. This (1) allows us to manage the combinatorial complexity caused by operating on a discrete number of objects, while (2) it also enables us to operate a robot in a closed-loop setting, where the action symbols are used to select policies for  solving sub-tasks.  

\subsection{Performing Actions}

\begin{figure*}[t]   
  \begin{center}
  \includegraphics[width=\linewidth]{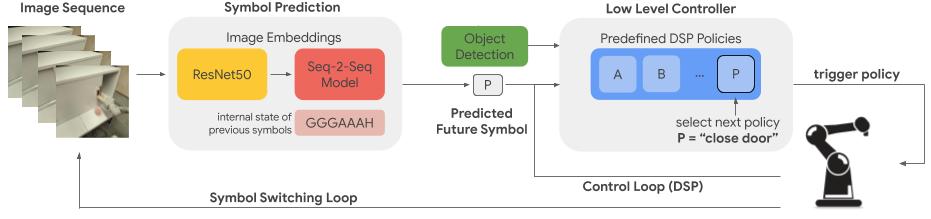} 
  \end{center}
  \vspace{-2mm}
  \caption{Pipeline of our framework: we obtain image embeddings of an incoming sequence of frames. A sequence of embeddings is used as the input of a sequence-to-sequence model that translates the embeddings to a sequence of action symbols. The sequence-to-sequence model is trained to predict the next action symbol based on a sequence of previous image embeddings. The predicted next action symbol is then passed to a low level controller that selects a corresponding policy to perform the action. Policies are reusable for different objects and are parameterized by the pose of individual objects that we identify through object detection.}
  \label{fig:pipeline}   
  \vspace{-5mm}
\end{figure*}

\ch{
To perform actions we model an action symbol as motion primitives~\cite{schaal05}. A motion primitive is a parameterized policy to perform an atomic action, such as grasping, placing, etc. Primitives can be used as building blocks that can be composed to represent tasks. 
Our symbol prediction network predicts sequences of action symbols from a history of image embeddings. This allows us to respond to changes in the environment, while the symbols provide structure to perform long-horizon planning. This is particularly useful if the next step of a task cannot be inferred from the current scene state. For example, if an object is hidden a system only operating on image embeddings may fail as the sought object is not available. Our approach instead also relies on the sequence of previous action symbols to infer the next step. 

We modeled each of our motion primitive as a dynamical systems policy (DSP)~\cite{Khansari16}, which can be trained from a few demonstrations. Given a target object pose, DSP drives the robot arm from its initial pose to the target pose while executing similar motions as provided in the demonstrations. In our setup we train each primitive based on five demonstrations captured through kinesthetic demonstrations. The input to each DSP primitive is the current object and arm end-effector pose, and the output is the next end-effector pose. Our robot is equipped with a perception system that performs object detection and classification~\cite{8237584} and provides the Cartesian pose of each object with respect to the robot frame. We use the object pose as parameter for the DSP primitives, which allows us to reuse primitives for multiple objects.  

Fig.~\ref{fig:pipeline} illustrates the overall architecture of our system. Once the sequential model determines the next action, the corresponding primitive is called with the poses of relevant objects and the robot starts executing the motion. Note that there are two loops in our system: (1) the DSP control loop which runs at 20Hz and is in charge of moving the arm to the target location, and (2) the symbol switching loop which runs at 2Hz and determines the next primitive that needs to be executed solely based on the stream of images. 
}

\subsection{Network Architecture and Training}
\label{label:architecture}

The sequence-to-sequence network is trained on sequences of image embeddings and action symbols. Instead of training on the full sequences, we train the network on sub-sequences of a specified sequence length (SL). Specifically, we experimented with the sequence lengths 10, 20, and 30 (Tab.~\ref{table:seq2seq_errors}). The sub-sequences are generated as `sliding windows' over an entire sequence. We train the model so as to translate sequences of image embeddings to predict a sequence of action symbols. However, the sequence of predicted action symbols are offset by $k$, where $k$ represents the number of steps we want to predict in the future. For our experiments we mostly relied on setting~$k=1$, which means that we only predict the action one step ahead in the future. The encoder takes the input frame embeddings and generates a state embedding vector from its last recurrent layer, which encodes the information of all input elements. The decoder then takes this state embedding and converts it back to action symbol sequences. The sequence-to-sequence model is trained with a latent dimension of 256 and usually converges after 50 epochs. We train the adapted ResNet50 network on single pairs of images and action symbols and randomly select these pairs from all sequences of our training data. The network is trained until it converges, which usually happens after no more then 200 epochs for our datasets. 
For both networks we did not specifically finetune the hyperparameters. 

\vspace{-2mm}
\section{DATASETS}
\label{lab:datasets}
\vspace{-2mm}

To validate the usefulness of a symbol-based action prediction model for manipulation tasks we defined two different datasets. The details for each are summarized in the Supplementary Material (SP), Tab.~\ref{table:dataset}. 

\textbf{Manipulation.} For this dataset we defined object manipulation tasks, where the goal is to put a ball in a cup. However, the cup and the ball can either be hidden in the cabinet or located somewhere in front of it. Depending on the object configuration, the task then becomes to first open the cabinet, to grasp the cup and the ball, to move them outside the cabinet, then to drop the ball into the cup, and to move the cup with the ball back into the cabinet. Finally, the cabinet door needs to be closed. The cabinet door has a handle and can be opened by a sliding mechanism. 
\begin{figure*}[t]   
  \begin{center}
  \includegraphics[width=0.95\linewidth]{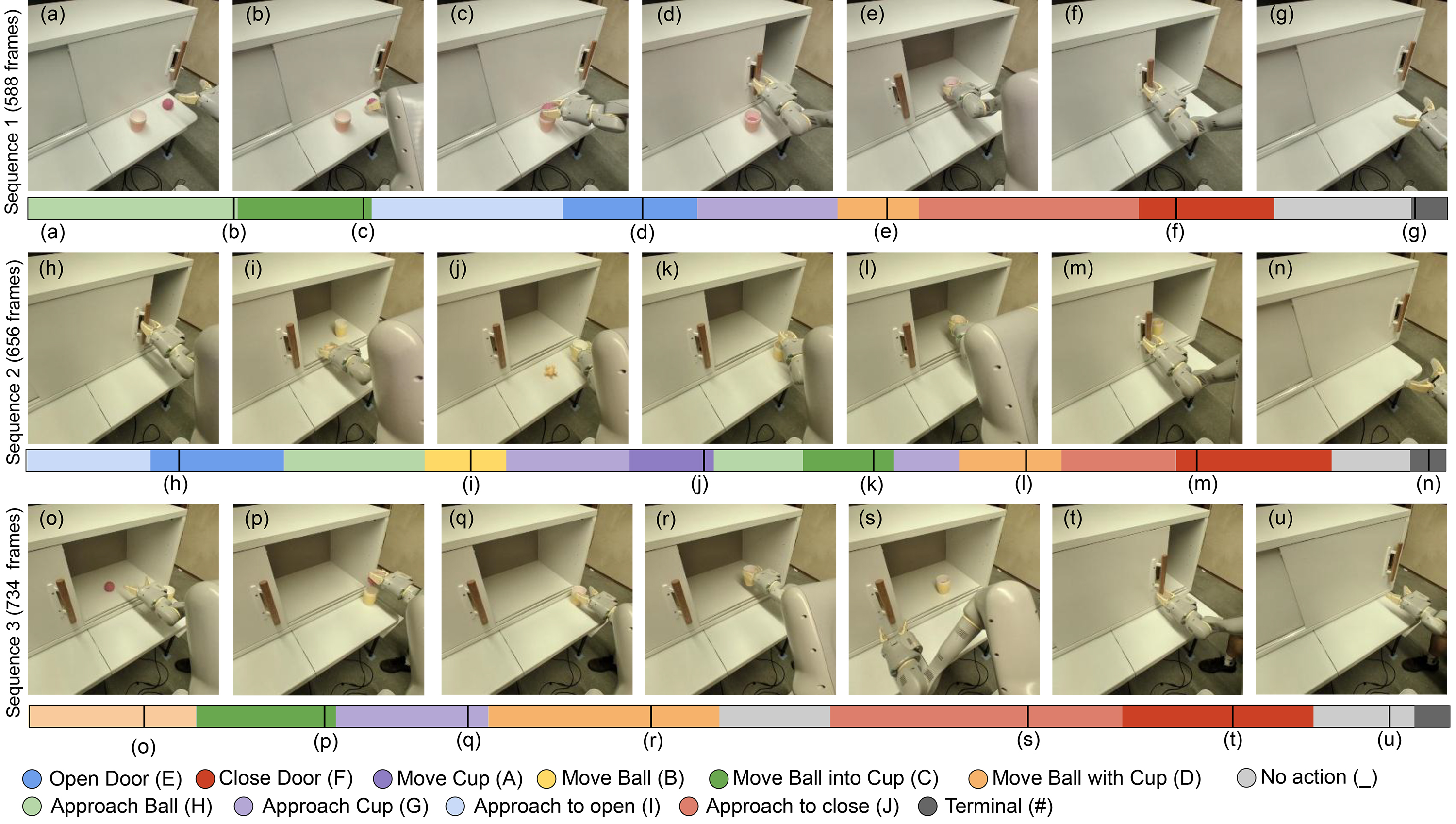} 
  \end{center}
  \vspace{-3mm}
  \caption{Manipulation of objects over long time horizons: the same task is performed in a different order and with different subsets of actions. Which actions are necessary to achieve the goal (cup in cabinet, closed cabinet) depends on the initial scene configuration. For the sequence (a)-(g) cup and ball are initially outside the cabinet. 
  For the sequence shown in (h)-(n) both objects (cup, ball) are inside the cabinet and the only possible action is to first open the cabinet. Finally, for the sequence (o)-(u) the cabinet door is initially open. 
  The bars underneath the images illustrate the structure of  task-relevant actions; black bars indicate where in the sequence the shown frames were taken from.}
  \vspace{-5mm}
  \label{fig:manipulation_seqs}  
  
\end{figure*}
Given this setting we define four different tasks: the easiest task is to just move a ball in a cup (Manipulation C). Here the model only needs to predict the correct order of two symbols (G: approach cup, C: move ball into cup). We then make this task gradually more complex by adding action symbols, e.g. to first move ball and cup out of the cabinet before putting the ball in the cup (Manipulation ABC), by moving the cup with the ball back to the cabinet after the other actions have been performed (Manipulation ABCD), and finally, to also open and close the door of the cabinet (Manipulation ABCDEF). Please note that for the tasks (ABC, ABCD, and ABCDEF), the order of the action symbols can vary. For example, it is possible to first move the cup or the ball outside the cabinet. However, some actions need to be executed before others: the door of the cabinet has to be open before the cup with the ball can be moved into the cabinet. 

A human operator places the objects into the scene and then performs one of the tasks with the robot arm, controlled by a tele-operation system (SP, Fig.~\ref{fig:teleop}). Each sequence consists of 130 - 890 frames and in total we captured 839 sequences. Across the sequences we define tasks with different levels of complexity, going from just moving the ball into the cup, to the full range of actions described above. We used a 80-10-10 split for training, validation, and test data. Possible actions are `move cup', `move ball', `move ball to cup', `move ball and cup', `open door', `close door', and the corresponding approach motions, e.g. `approach cup', `approach to open', etc. (SP, Tab.~\ref{table:action_symbols}). Frames of the captured sequences are manually labeled with the respective action symbols.

\textbf{Block Stacking.} For the second dataset we define a stacking task of five uniquely shaped blocks (similar to Tetris blocks) of different colors. The goal is to generate an object arrangement of two stacks and one single block. For each run the blocks are placed randomly on a table. An operator then takes the blocks and stacks them into the specified configuration. The order in which the objects are stacked is not defined. However, to generate the target configuration of objects only certain action sequences are plausible. While the blue block can be placed any time during the task, the green and the pink block require to first place the yellow and red block respectively. The example in Fig.~\ref{fig:blocks} (SP) shows three initial and the target configuration of objects. We captured 289 sequences (150 - 450 frames) of this stacking task and used a 80-10-10 split for training, validation, and test data. 

As the manual labeling of frames with action symbols is a laborious task we used IMU sensors embedded in each block to automatically obtain action symbols (SP, Fig.~\ref{fig:imu}). This is based on the observation that to understand the structure of a task we only need to know the order in which the objects where set in motion. To compute an action symbol we track the IMU signal for each individual block by fusing the acceleration and orientation signal~\cite{Madgwick2010AnEO} of a single sensor. We then set a flag for whether the corresponding object is set in motion. As the task requires to stack the objects, multiple object motions may be detected. However, as we are not interested in tracking multiple objects at the same time, we only consider the object with the larger changes of acceleration and orientation. We use Bluetooth enabled IMU boards to automatically obtain action symbols for multiple blocks simultaneously. Based on the available objects the possible actions for this task are to move any of the available blocks (blue, red, yellow, pink, green) and `no action' (SP, Tab.~\ref{table:dataset}). 
\vspace{-3mm}
\section{EXPERIMENTS AND RESULTS}
\vspace{-3mm}
In this section we evaluate the performance of our framework on sequence prediction for manipulation tasks. The goal is to predict sequences of action symbols that describe the sequential structure of a task and thereby allow an agent to execute a task in the correct order. The sequence of action symbols is predicted based on a sequence of frame embeddings. This allows us to reliably predict the next action based on the current state of the scene.  

\vspace{-1mm}
\subsection{Sequence Translation and Prediction} 
\vspace{-1mm}

To evaluate the quality of the predicted sequences we use our sequence-to-sequence model to predict the next action symbol based on a sequence of image embeddings and compare the result of a whole predicted sequence with the ground truth sequence. We then measure three different errors over these sequences. \ch{In Tab.~\ref{table:seq2seq_errors} we report results for all metrics and for sequence lengths (SL) of 10, 20, and 30. First, the symbol-to-symbol error measures the overall accuracy of the predicted sequences; each symbol is compared to its corresponding symbol in the ground truth sequence. 
This error metric provides a way to measure the overall accuracy but it does not account for the impact a wrongly predicted symbol may have for executing a sequence; i.e. if only a single symbol is predicted wrongly, executing the task may fail. Example sequences for both task seutps are shown in Figures~\ref{fig:manipulation_seqs} and \ref{fig:sequence_example}, SP.}

Therefore, we additionally compute the error of predicting the correct sequential structure of actions. For this we again predict an action symbol for each frame in a sequence and then shorten the sequence of symbols to its compact representation; i.e. when the same symbol was predicted repeatedly for consecutive frames we only use the symbol once (similar to Huffman coding). We then compare the predicted sequence with the ground truth sequence in its compact encoding and measure an error when there are any differences in the symbol patterns. The results for the sequences of our datasets are shown in Tab.~\ref{table:seq2seq_errors} (Structure). An example of shortened sequences is shown in the SP, Tab.~\ref{table:action_symbols} (Compact Example). When computing the structure error on the compact representation, irregularities in the sequential structure are accounted for. A single change of a symbol would create a different compact encoding and the sequence would be labeled as wrongly predicted. Finally, we use Levenshtein distance as a more common way to compare symbol sequences (Tab.~\ref{table:seq2seq_errors}, Edit Dist). Here the error is measured as the number of edit operations necessary to convert the predicted sequence into the ground truth sequence. The resulting number of edits is normalized by the number of symbols of the ground truth sequence. 

 Additionally, we compare the results of the sequence-to-sequence model with a many-to-many LSTM and soft attention weighted annotation~\cite{DBLP:journals/corr/BahdanauCB14} (Tab.~\ref{table:seq2seq_errors}). The LSTM consists of a single layer with a latent dimension of 256.  Compared to the sequence-to-sequence model the LSTM performs well on the symbol and edit distance metrics, which means that the overall distance of ground truth and predicted sequences is small. However, it performs with significantly less accuracy on predicting the structure of tasks.

\begin{table*}[t]
\begin{center}

\caption{Sequence Prediction Errors (Symbol, Structure, Edit Distance) for Seq2Seq and LSTM.}
\scalebox{0.76}{
\label{table:seq2seq_errors}
\begin{tabular}{l|l|c|c|c|c|c|c|c|c|c}
 &             & \multicolumn{3}{c|}{Symbol}                           & \multicolumn{3}{c|}{Structure}      & \multicolumn{3}{c}{Edit Distance}      \\
 & Method                  & SL10 & SL20 & SL30                                      & SL10 & SL20 & SL30                    & SL10 & SL20 & SL30 \\
\hline
\hline
\multirow{3}{*}{\rotatebox[origin=c]{90}{\parbox[c]{1.9cm}{\centering SEQ2SEQ}}} & Manipulation - ABCDEF  
                                             & 7.84\% & {6.62\%} & 6.23\%     & 21.42\%  & {18.02\%}  & 28.14\%    & 6.52\% & {5.82\%} & 6.62\%     \\ 
                     & Manipulation - ABCD   & 7.33\% & {5.98\%} & 6.71\%     & 20.02\%  & {14.26\%}  & 15.38\%    & 5.85\% & 5.21\% & { 5.26\%}       \\
                     & Manipulation - ABC    & {5.46\%} & 5.38\% & 5.23\%     & 16.01\%  & 11.53\%  & {10.14\%}    & 5.46\% & {4.86\%} & 5.74\%       \\
                     & Manipulation - C      & 3.23\%  & {4.01\%}  & 4.57\%   & 15.38\%   & {5.03\%}  & 5.08\%     & 4.79\% & {4.02\%} & 5.12\%       \\
\cline{2-11}                     
                     & Blocks                & 8.82\% & {7.65\%} & 8.15\%        & 29.28\%  & {13.03\%}  & 26.54\%      & {7.15\%} & 8.46\% & 9.03\%       \\
                     
\hline                     
\multirow{3}{*}{\rotatebox[origin=c]{90}{\parbox[c]{1.9cm}{\centering LSTM}}} & Manipulation - ABCDEF  
                                             & 11.06\% & 9.41\% & {7.90\%}     & 57.69\%  & 45.23\%  & {34.61\%}      & 10.48\% & 8.29\% & {7.37\%}     \\ 
                     & Manipulation - ABCD   & 10.56\% & 8.70\% & {7.85\%}     & 47.26\%  & 44.76\%  & {28.98\%}      & 9.38\% & 7.81\% & {6.99\%}       \\
                     & Manipulation - ABC    & 10.03\% & 7.82\% & {7.69\%}     & 35.01\%  & 40.58\%  & {27.19}\%      & 9.05\% & 6.90\% & {6.27\%}       \\
                     & Manipulation - C      & 8.80\% & 7.39\%  & {6.92\%}     & {25.92\%}  & 29.62\%  & 25.93\%      & 7.27\% & 6.81\% & {5.92\%}       \\
\cline{2-11}                     
                     & Blocks                & 12.44\% & {10.31\%} & 10.85\%     & 38.57\%  & 51.14\%  & {44.28\%}      & 10.48\% & {9.03\%} & 9.37\%       \\
\hline

\end{tabular}
}
\end{center}
\vspace{-5mm}
\end{table*}

\vspace{-2mm}
\subsection{Robotic Manipulation}
\vspace{-1mm}

To test our framework in a realistic setting, we use a real robot to perform the manipulation task based on predicted motion primitives. Our robot is a 7-axis robotic arm equipped with an on-board camera. 

\ch{We setup a scene with randomly placed objects (cup, ball, open/closed door). We then generate image embeddings of the incoming camera frames. After obtaining \textit{sequence length (SL)} number of frames (e.g. 10, 20 or 30) the sequence-to-sequence model starts predicting the next action. With additional incoming frames the model keeps predicting the same symbol if no changes in the scene occur. The predicted symbol is then passed to the low-level controller and the robot is set in motion. If the scene changes, e.g. if the robot starts moving from its default position to the cup, the model starts predicting new action symbols.} A newly predicted action symbol is then pushed to a queue if it is different from the previous symbol in the queue. The robot takes the next action symbol from the queue, performs object detection to obtain the object poses, and runs the motion primitive corresponding to the selected symbol. While the robot is performing the action, the sequence prediction network predicts further action symbols that are added to the queue. When the robot is done with an action, it proceeds to the next action symbol and continues the task. Once all actions are completed successfully, the sequence model predicts the terminal symbol (\#) and the robot moves back to its default state.

\setlength{\intextsep}{5pt}%
\setlength{\columnsep}{10pt}%
\begin{wraptable}{r}{0.55\textwidth}
\vspace{-6mm}
\begin{center}
\caption{Robot Task Execution Accuracy.}
\scalebox{0.69}{
\label{table:robot_success}
\begin{tabular}{l|c|c|c|c}
Method & \#Success & \#Recovered & \#Failure  & Accuracy            \\
\hline
\hline
Manipulation ABCDEF  & 8  & 8   & 4   & 80.0\%       \\
Manipulation ABCD & 13 & 3  & 4   & 80.0\%      \\
Manipulation ABC  & 12  & 5   & 3   & 85.0\%        \\
Manipulation C    & 17 & 1  & 2  & 90.0\%       \\
\end{tabular}
}
\end{center}
\vspace{-2mm}
\end{wraptable}
To evaluate how well the robot is able to perform the tasks we measure how often it was able to successfully reach the goal state of a given task. We ran every task 20 times and counted the number of successes and failures. Depending on the scene setup, some predicted symbol sequences are implausible and their execution fails; we count these as failures. However, as our model relies on images embeddings to predict the next action, for some of these sequences the model can recover. Here the model may predict a wrong action symbol, but it then recovers and eventually predicts the correct sequence of actions to arrive in the goal state; we consider these sequences successes. Successful, recovered, and failed task completions are reported in Tab.~\ref{table:robot_success}.

\begin{figure}[t]
\begin{center}
  \includegraphics[width=\linewidth]{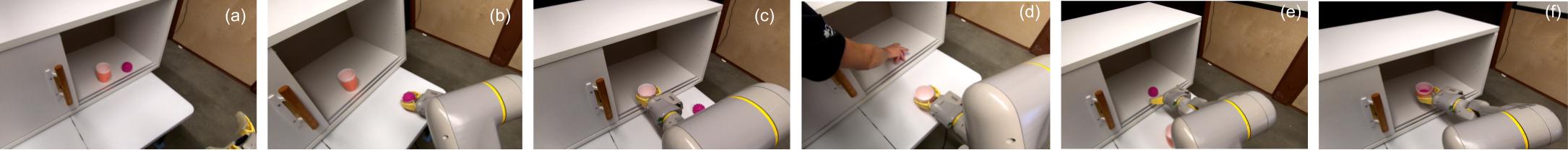} 
  \end{center}
  \vspace{-4mm}
  \caption{Dynamic interaction: the robot is executing the task of putting a ball in a cup. From an initial object arrangement (a) it first fetches a ball~(b), then continues to get the cup (c). A human operator then puts the ball back inside the cabinet (d) and the robot dynamically adapts by changing the plan to again fetch the ball (e). Finally, it  completes the task by putting the ball in the cup and both objects back into the cabinet (f). 
  }
  \label{fig:dynamic_interaction}   
\vspace{-5mm}
\end{figure}

\vspace{-2mm}
\subsection{Closed-loop Response} 
\vspace{-2mm}

\begin{figure}[h]   
  \begin{center}
  \includegraphics[width=\linewidth]{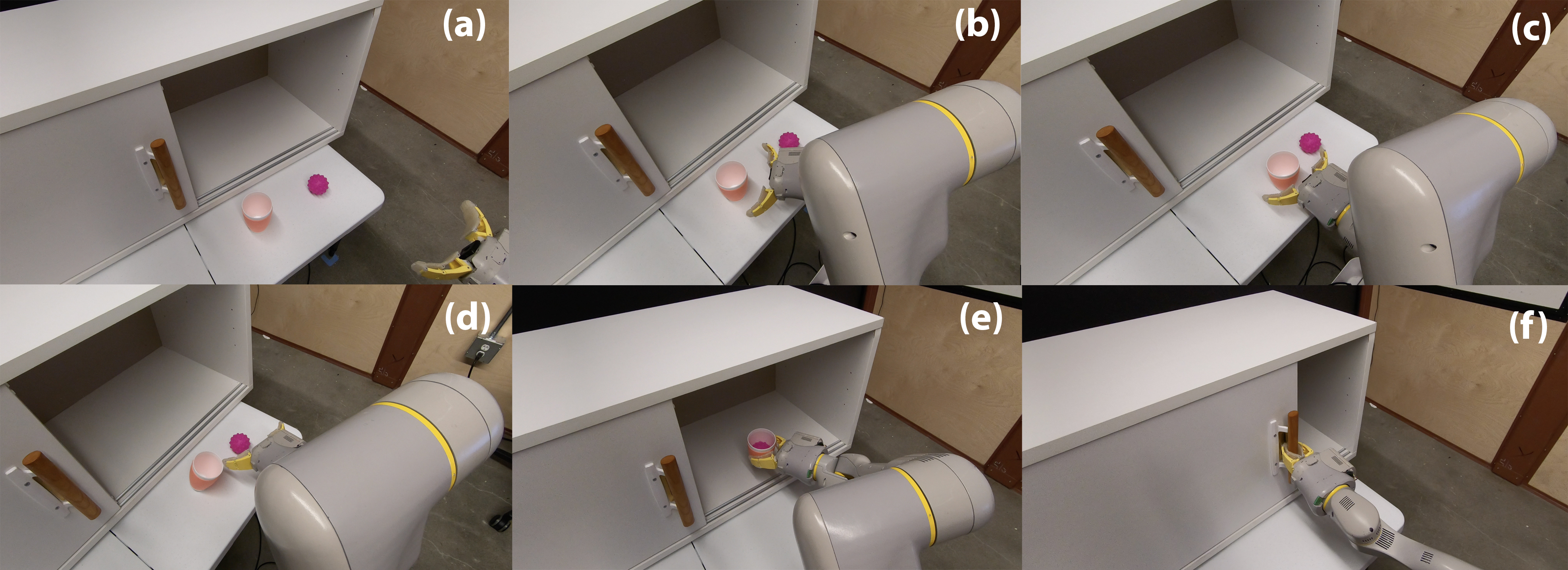} 
  \end{center}
  \vspace{-3mm}
  \caption{Recovering from wrong predictions: depending on the scene setup predicted symbols can be implausible or their execution may fail. For the scene shown in (a) a plausible action would be to put the ball in the cup, but the system predicts to first move the cup (b) \& (c). In (d) the gripper is accidentally moving the cup while the arm reaches for the ball. In both cases our system can recover from these failure states as we rely on image embeddings for predicting action symbols. Eventually the robot is then able complete the task (e), (f).}
  \label{fig:recover}   
\end{figure}

As shown in Fig.~\ref{fig:pipeline}, our model works in a closed-loop setting -- image embeddings are translated to action symbols. Therefore, a robot can interactively adapt to changes in the environment to successfully finish a task. In Fig.~\ref{fig:dynamic_interaction} we show the results of dynamic scene changes. While the robot is working on a task, we interfere with the scene and move objects around. \ch{ In the example shown in Fig.~\ref{fig:dynamic_interaction} (d), the user puts the ball back into the cabinet while the robot is placing the cup in front of the cabinet. The robot visually detects this change and retrieves the ball after placing the cup. Here the adaptation is happening at two levels: first the sequence model detects the ball misplacement through the image embeddings and again triggers the `grasp ball primitive`, then the primitive receives the new ball position through the perception system and thus adapts its motion in order to grasp the ball. 
In Fig.~\ref{fig:recover} (SP) we show how our pipeline is able to recover from wrongly predicted actions 
and in the accompanying video we show several live-action captures of our system to showcase the capabilities of our framework.
}

\vspace{-2mm}
\section{CONCLUSION}
\vspace{-3mm}
We have introduced a framework for translating sequences of action symbols from image embeddings. Each symbol represents a task-relevant action that is necessary to accomplish a sub-task. We have shown that symbols serve as a lightweight and abstract representation that not only enables using sequential models -- known from natural language processing -- for the efficient learning of task structure, but also to organize the execution of tasks with real robots. Learning to translate image embeddings to action symbols allows us to execute tasks in closed-loop settings, which enables robots to adapt to changing object configuration and scenes. We have demonstrated the usefulness of our framework on two different datasets and evaluated our approach on two model architectures. One limitation of our current setup is that we rely on ground truth labels for actions in the observed sequences. We addressed this limitation by automatically capturing  action symbols through IMU sensors embedded in individual objects. Automatically obtaining these action labels at scale is not trivial and an interesting avenue for future research. \chch{Another limitation is that we did not investigate to condition the prediction of our model on the task, e.g. by providing a goal state. While this limits our method in some scenarios, it seems plausible that a representation of the goal state can be provided as an additional embedding to disambiguate the task or to enable learning multiple tasks with the same model.} Finally, in many situations it would be important to understand the sequential structure of tasks over even longer time-horizons. Here it seems interesting to explore other model architectures for sequence translation from image embeddings.


{
\bibliography{main}  
}

\newpage
\section*{Appendix}

\subsection*{Encoder-Decoder Architecture}

\begin{figure}[h]
  \begin{center}
  \includegraphics[width=0.8\linewidth]{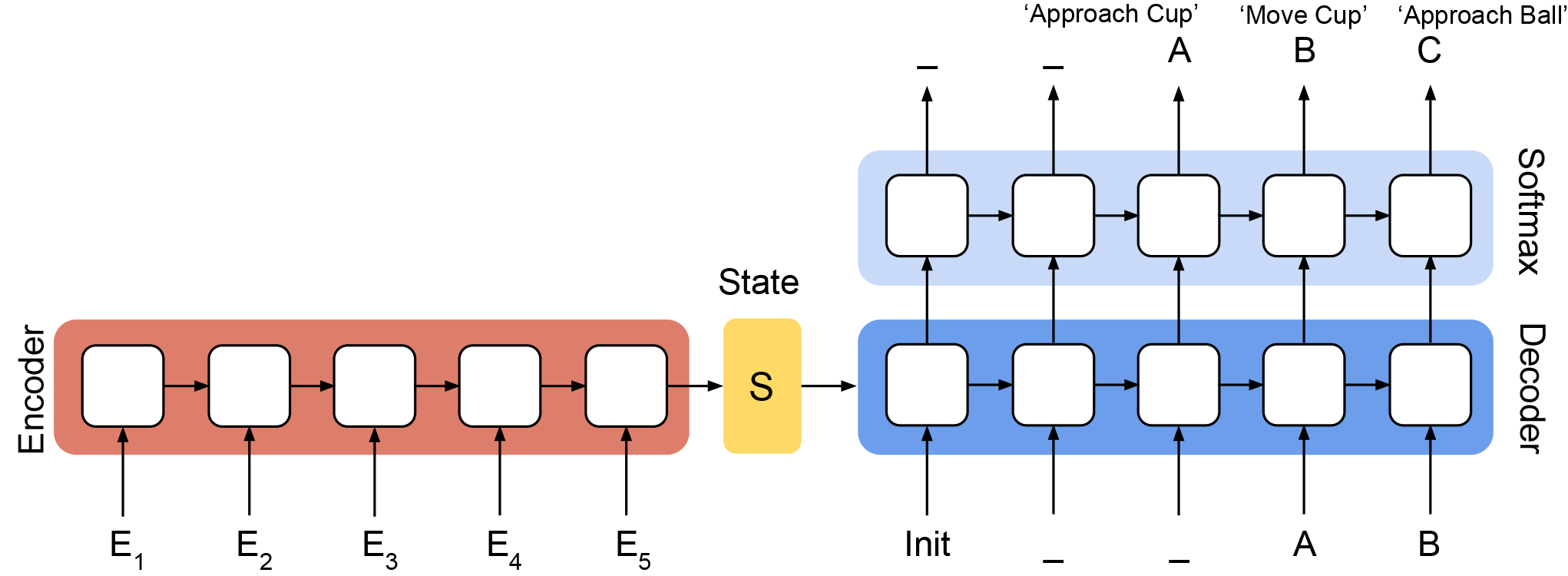} 
  \end{center}
  \vspace{-3mm}
  \caption{Encoder-Decoder architecture of a sequence-to-sequence model: image embeddings ($E_i$) serve as input for the encoder. The encoder converts the embedding to a state vector (S). The decoder uses the state vector and a symbol sequence to predict the output sequence one symbol ahead of the current time step.}
  \label{fig:encoder_decoder}   
\end{figure}

We use a sequence-to-sequence model~\cite{Sutskever:2014:SSL:2969033.2969173} to translate image embeddings to action symbols. The model is setup to predict a future action symbol given a history of image embeddings as illustrated in Figure~\ref{fig:encoder_decoder}.

\vspace{-2mm} 
\subsection*{Action Symbols}
For both tasks (manipulation, block stacking) task-relevant actions are assigned an action symbol as shown in Tab.~\ref{table:action_symbols}. Please note that for the block stacking tasks action symbols are automatically obtained based on IMU sensors as described in Section~\ref{lab:datasets} and shown in Fig.~\ref{fig:imu}. We differentiate between class symbols (Manipulation) and instance symbols (Block Stacking). For the manipulation task we represent objects of the same class (e.g. cups) with the same symbol as we only operate on one object of a specific class at a time. If the task requires to simultaneously operate on multiple objects of the same class (e.g. blocks) symbols for individual instances are used.

\begin{table}[h]
\begin{center}
\vspace{2mm} 
\caption{Action symbols and their meaning for both datasets.}
\scalebox{0.77}{
\label{table:action_symbols}
\begin{tabular}{l|c|c|l}
Dataset & Action Integer & Action Symbol & Meaning\\
\hline
\hline
Manipulation    & 0  & A & Move cup      \\ 
                & 1  & B & Move ball      \\ 
                & 2  & C & Move ball into cup      \\ 
                & 3  & D & Move ball and cup      \\ 
                & 4  & E & Open door      \\ 
                & 5  & F & Close door      \\ 
                & 6  & G & Approach cup      \\ 
                & 7  & H & Approach ball      \\ 
                & 8  & I & Approach to open      \\ 
                & 9  & J & Approach to close      \\ 
                & 10 & \_ & No action      \\
                & 11 & \# & Terminal/Done      \\
\cline{2-4}                
& \multicolumn{3}{l}{Compact Example: EBACDF\_} \\
\hline
Block Stacking  & 0 & B & Move Blue      \\ 
                & 1 & R & Move Red      \\ 
                & 3 & Y & Move Yellow      \\ 
                & 3 & G & Move Green      \\ 
                & 4 & P & Move Pink      \\ 
                & 5 & \_ & No action      \\ 
                
\cline{2-4}
& \multicolumn{3}{l}{Compact Example: \_Y\_B\_G\_R\_P\_} \\
\hline
\end{tabular}
}
\end{center}
\vspace{-4mm}
\end{table}

\subsection*{Dataset Details}

\begin{table}[h]
\begin{center}
\caption{Details of our two datasets.}
\scalebox{1.0}{
\label{table:dataset}
\begin{tabular}{l|c|c|c|c|c}
Dataset & \#Tasks & \#Sequences & \#Symbols & \#Frames & Sequence Length\\
\hline
\hline
Manipulation & 4 & 791 & 7 & 228K & 90-600    \\ 
Blocks       & 1 & 287 & 6 & 97K & 136 - 445 \\ 
\end{tabular}
}
\end{center}
\end{table}

\subsection*{Block Stacking Task}
\begin{figure}[h]   
  \begin{center}
  \includegraphics[width=\linewidth]{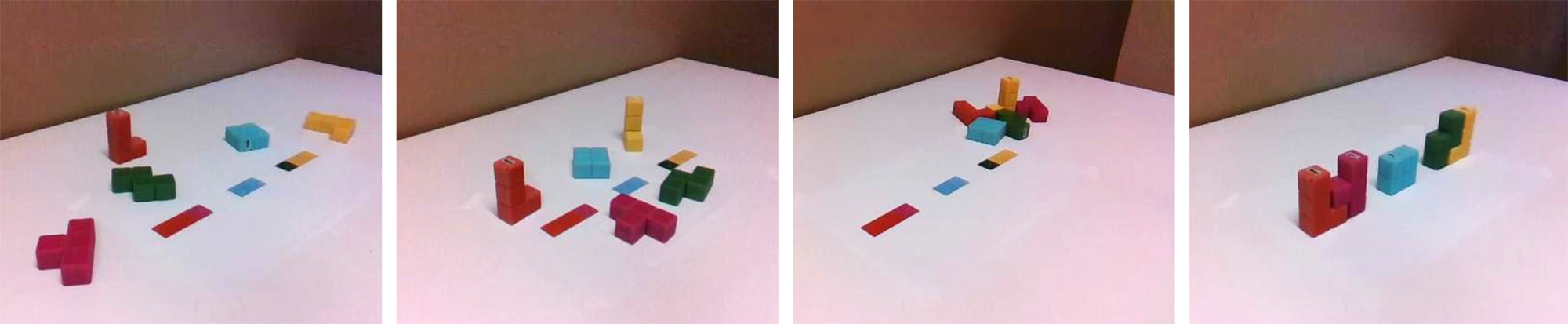} 
  \end{center}
  \vspace{-3mm}
  \caption{For the block stacking task we randomly place the five blocks in the scene. The goal is to then move them  to a stacked configuration. From left to right: three initial configurations of objects and the final configuration.}
  \label{fig:blocks}   
\end{figure}

\begin{figure}[h]   
  \begin{center}
  \includegraphics[width=\linewidth]{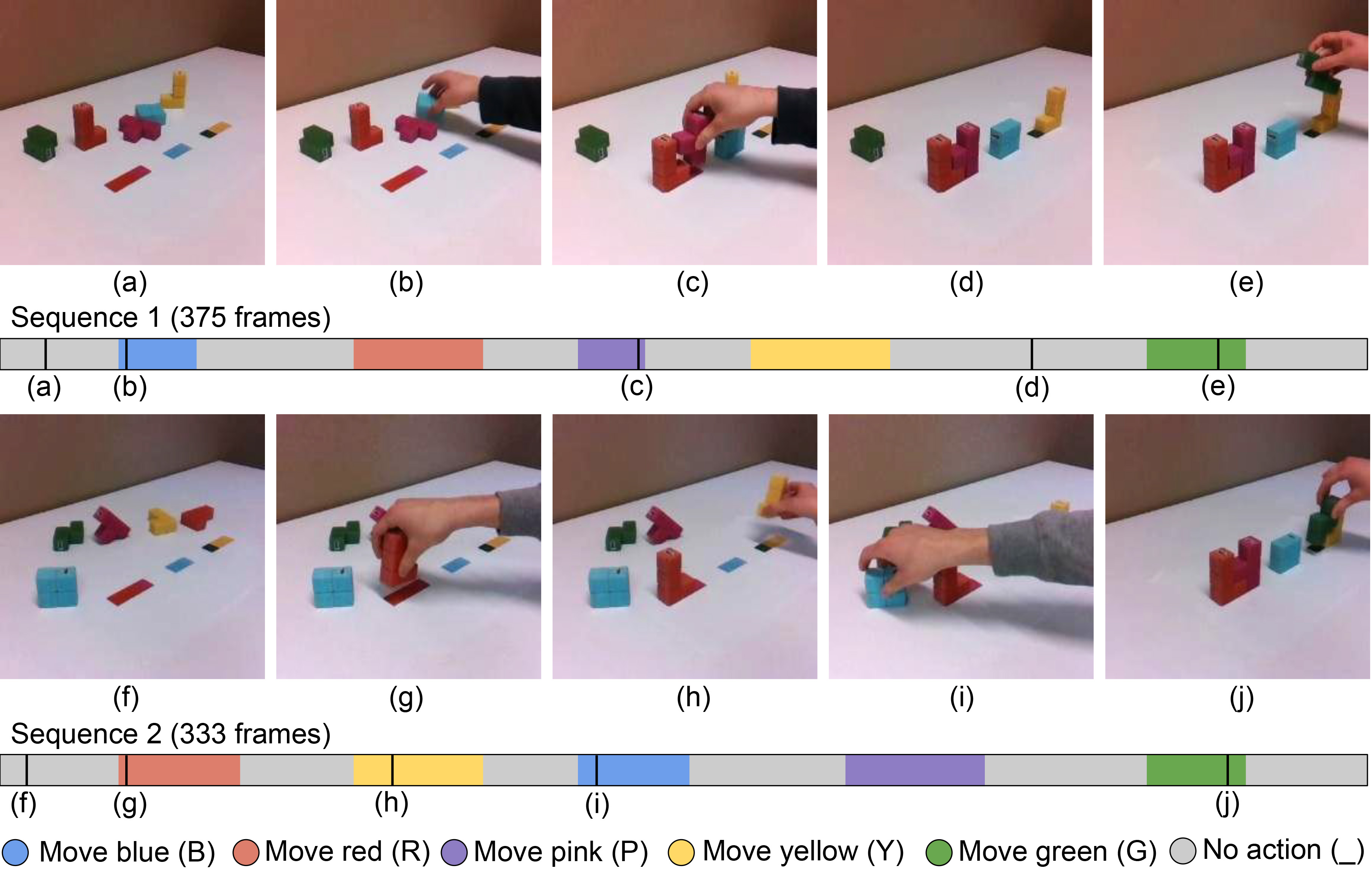} 
  \end{center}
  \vspace{-3mm}
  \caption{Two example sequences of a blocks stacking task: we associate action symbols to every frame of a sequence (a-e) and (f-j). An action symbol represents the action in an abstract way (e.g. move red block). For the frames (a-e) and (f-j) we show where in the sequence this action was performed. The order of actions can vary, but sequential dependencies exist.}
  \label{fig:sequence_example}   
  \vspace{-2mm}
\end{figure}

\vspace{-1mm}
Figures~\ref{fig:blocks} and~\ref{fig:sequence_example} show our block stacking setup and two variations of preforming this task. 


\subsection*{Capturing Data with a Tele-op System}
\begin{figure}[h]   
\begin{center}
  \includegraphics[width=1.0\linewidth]{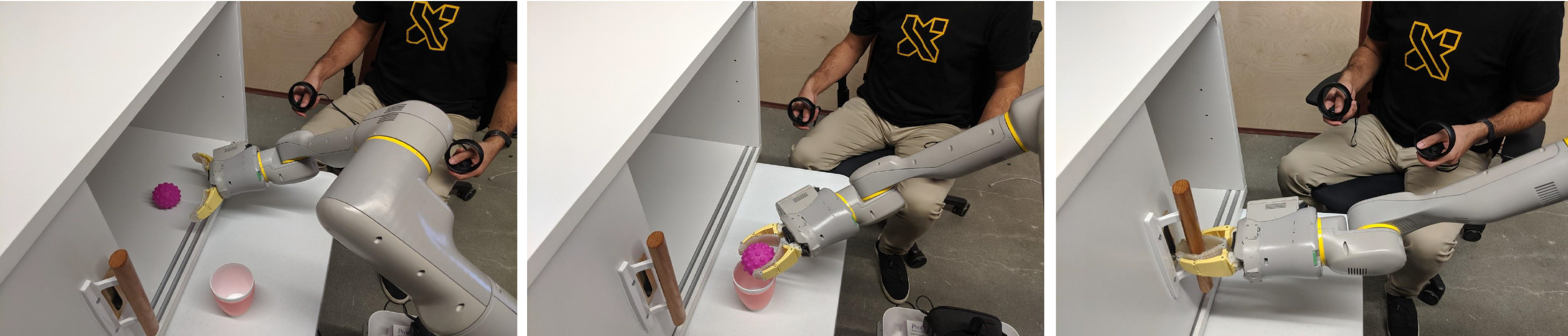} 
  \end{center}
  \vspace{-3mm}
  \caption{Human operator performing a task of grasping a ball, putting it into a cup, and closing a cabinet door, performed with a teleop system.}
  \label{fig:teleop}     
\end{figure}

\vspace{-2mm}
To train a model on image embeddings that can be used with real robots and scenes, we relied on a tele-op system. This setup allows us to capture example sequences as training data, while observing the scene from the perspective of the robot. Figure~\ref{fig:teleop} shows examples of our tele-op system and the perspective of the robot while capture data for our robotic manipulation task.

\subsection*{Automatically Capturing Action Symbols}

\begin{figure}[h!]   
  \begin{center}
  \includegraphics[width=\linewidth]{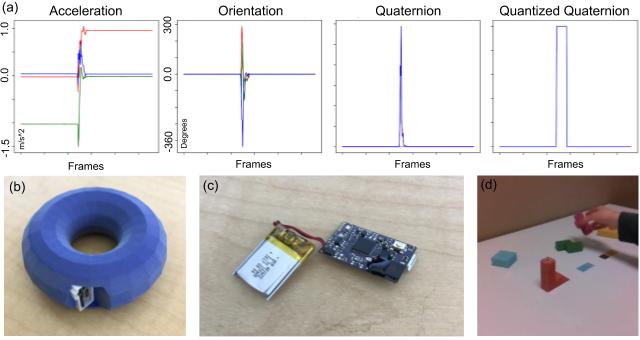} 
  \end{center}
  \vspace{-3mm}
  \caption{ 
  Each IMU sensor provides us with an acceleration and orientation signal that can be fused into a quaternion \cite{Madgwick2010AnEO} (a). We quantize the quaternion and use the resulting signal to obtain an action symbol for the video frames where the respective object is set in motion. An example object along with the used IMU sensor board is shown in (b) and (c). IMU sensors allow us to track the motion of individual objects (d).}
  \label{fig:imu}   
  \vspace{3mm}
\end{figure}

We embed a IMU sensors\footnote{https://mbientlab.com/} in each block to automatically obtain action symbols. Figure~\ref{fig:imu} shows an example of a captured IMU signal along with an example object and the used IMU sensor board.

\subsection*{Closed-loop Response: Recovering from Failure}

\end{document}